\def\year{2019}\relax
\documentclass[letterpaper]{article} %DO NOT CHANGE THIS
\usepackage{aaai19}  %Required
\usepackage{times}  %Required
\usepackage{helvet}  %Required
\usepackage{courier}  %Required
\usepackage{url}  %Required
\usepackage{graphicx}  %Required
\usepackage{booktabs} % For formal tables
\usepackage{xcolor}
\usepackage{color}
\usepackage{colortbl}
\usepackage{soul}
\usepackage{amsfonts}
\usepackage[small]{caption}
\usepackage{balance} 
\usepackage[linesnumbered,ruled,vlined]{algorithm2e}
\usepackage[flushleft]{threeparttable}
\usepackage{tablefootnote}
\usepackage{enumitem}
\usepackage{graphics}
\usepackage{subcaption}
\usepackage{amsmath}
\usepackage{comment}
\usepackage{pifont}% http://ctan.org/pkg/pifont
\newcommand*{\affaddr}[1]{#1} % No op here. Customize it for different styles.

\newcommand*{\email}[1]{\texttt{#1}}
\usepackage{array}
\usepackage{float}
\newcolumntype{P}[1]{>{\centering\arraybackslash}p{#1}}
\newcolumntype{M}[1]{>{\centering\arraybackslash}m{#1}}

\begin{document}
\title{ATP: Directed Graph Embedding with Asymmetric Transitivity Preservation}
\author{%
\small Jiankai Sun,
Bortik Bandyopadhyay, Armin Bashizade, Jiongqian Liang, P. Sadayappan, and Srinivasan Parthasarathy\\
\affaddr{\small  Department of Computer Science and Engineering, The Ohio State University, USA}\\
\email{\footnotesize \{sun.1306,bandyopadhyay.14,bashizade.1,liang.420,sadayappan.1\}.osu.edu,srini@cse.ohio-state.edu}\\
}
% \author{%
% \small Jiankai Sun\affmark[1],
% Bortik Bandyopadhyay\affmark[1], Armin Bashizade\affmark[1], Jiongqian Liang\affmark[2], P. Sadayappan\affmark[1], and Srinivasan Parthasarathy\affmark[1]\\
% \affaddr{\small \affmark[1] Department of Computer Science and Engineering, The Ohio State University, USA}\\
% \affaddr{\small \affmark[2] Google, Mountain View, USA}\\
% \email{\affmark[1] \footnotesize \{sun.1306,bandyopadhyay.14,bashizade.1,sadayappan.1\}.osu.edu,srini@cse.ohio-state.edu}\\
% \email{\affmark[2] \footnotesize liang.albert@outlook.com}
% }
\maketitle
\begin{abstract}
Directed graphs have been widely used in Community Question Answering services (CQAs) to model asymmetric relationships among different types of nodes in CQA graphs, e.g., question, answer, user. Asymmetric transitivity is an essential property of directed graphs, since it can play an important role in downstream graph inference and analysis. Question difficulty and user expertise follow the characteristic of asymmetric transitivity. 
Maintaining such properties, while reducing the graph to a lower dimensional vector embedding space, has been the focus of much recent research. 
In this paper, we tackle the challenge of directed graph embedding with asymmetric transitivity preservation and then leverage the proposed embedding method to solve a fundamental task in CQAs: how to appropriately route and assign newly posted questions to users with the suitable expertise and interest in CQAs. The technique incorporates graph hierarchy and reachability information naturally by relying on a non-linear transformation that operates on the core reachability and implicit hierarchy within such graphs. Subsequently, the methodology levers a factorization-based approach to generate two embedding vectors for each node within the graph, to capture the asymmetric transitivity. Extensive experiments show that our framework consistently and significantly outperforms the state-of-the-art baselines on three diverse real-world tasks: link prediction, and question difficulty estimation and expert finding in online forums like Stack Exchange. Particularly, our framework can support inductive embedding learning for newly posted questions (unseen nodes during training), and therefore can properly route and assign these kinds of questions to experts in CQAs.
\end{abstract}

\section{Introduction}
\label{sec:introduction}

Community Question Answering services (CQAs) such as Stack Exchange and Yahoo! Answers are examples of social media sites, with their usage being examples of an important type of computer supported cooperative work in practice. In recent years, the usage of CQAs
has seen a dramatic increase in both the frequency of questions posted and general user activity.
This, in turn, has given rise to several interesting problems ranging from expertise estimation to question difficulty estimation, and from automated question routing to incentive mechanism design on such CQAs ~\cite{Fang2016QuestionAnswering,QDEE2018}. %Recently, Sun et al.~\cite{QDEE2018} observed that users typically gain expertise across multiple interactions with the CQA and tend to ask more difficult questions within the same domain over time, which is referred to {\em Expertise Gain Assumption} (EGA). They leveraged EGA and added additional edges between questions asked by the same user to the previous competition graph~\cite{Liu2011,liu2013question,wang2014} to combat the sparseness problem. They then proposed QDEE to lever social agony~\cite{Tatti2014,tatti2015} to infer graph hierarchy and assign each node a scalar value to represent where it stands in the competition graph. %The scalar value assigned to each node represents corresponding node's hierarchy information in the competition graph.
Previous work~\cite{wang2014,QDEE2018} proposed to assign a scalar value to represent question difficulty (and user expertise). %infer graph hierarchy and assign each node (question, user) a scalar value to represent where it stands in the graph.
%The assigned scalar value of question nodes are question difficulty scores, and the assigned scalar value of user nodes are user expertise scores.
However, question difficulty and user expertise can vary in different topics. In Stack Exchange sites, users are required to use tags (a tag is a word or phrase) to describe the topic(s) of the question \footnote{\url{https://stackoverflow.com/help/tagging}}. Each question can be assigned multi-tags to represent its most relevant topics. For example, in our experiments, the average number of tags per question is $2.82$ and $2.96$ in Stack Exchange site Apple and Physics respectively. Hence a solely scalar value to represent question difficulty level or user expertise is not thorough.

Some graph embedding methods~\cite{Fang2016QuestionAnswering,zhaoexpert2016,zhao2017community} are then proposed to address the above limitation. 
The problem of graph embedding seeks to represent vertices of a graph in a low-dimensional vector space in which meaningful semantic, relational and structural information conveyed by the graph can be accurately captured ~\cite{Ma2018WSDM}. %Classic vector-based machine learning techniques can leverage such vectors as features for many tasks such as link prediction, multi-label classification, clustering and vertex recommendation ~\cite{Harp2017}. 
Recently, one has seen a surge of interest in developing such methods including ones for learning such representations for directed graphs (while preserving important properties) ~\cite{Ou2016KDDAsymmetric}, which is the focus of our research. A property of singular importance within a directed graph is asymmetric transitivity, which plays a very important role in tasks of graph inference and analysis ~\cite{Ou2016KDDAsymmetric}. %\cite{Ou2016KDDAsymmetric,SOME2006,SimRank2002,Katz1953}.  %For example, one important category of input graph for embedding is built from Community Question Answering services (CQAs) such as Stack Overflow and Quora, which are Internet-based crowdsourcing services that enable users to post questions on websites, which are then answered by other users. In Stack Overflow, about 4.8 million questions have not been answered \footnote{\url{https://stackexchange.com/sites#}}. CQAs graph embedding can be leveraged to tackle a fundamental challenge in crowdsourcing: how to appropriately route and assign newly posted questions to users with the suitable expertise and interest in CQAs. This has great potential to increase question answering quality, satisfaction, helping all the users to improve their expertise (and not just top-level experts), and reduce inefficiency in real-time crowdsourcing of answers. 
Question difficulty and user expertise follow the characteristic of asymmetric transitivity. For example, given a question $q_1$ is easier than $q_2$ and $q_2$ is easier than $q_3$, we can infer that $q_1$ is easier than $q_3$ easily. It happens to estimating user expertise too. We can infer that $u_1$ has more expertise than $u_3$ based on the fact that $u_1$ has more expertise than $u_2$ and $u_2$ has more expertise than $u_3$ in a specific domain. In this paper, we tackle the challenge of directed graph embedding with asymmetric transitivity preservation and then leverage the proposed embedding method to solve a fundamental task in CQAs: how to appropriately route and assign newly posted questions to users with the suitable expertise and interest in CQAs. %For instance, in Stack Overflow, an Internet-based crowdsourcing services that enable users to post questions on websites, about 4.8 million questions have not been answered \footnote{\url{https://stackexchange.com/sites#}}.

\begin{figure}[t!]
    \centering
    \includegraphics[width=0.475\textwidth]{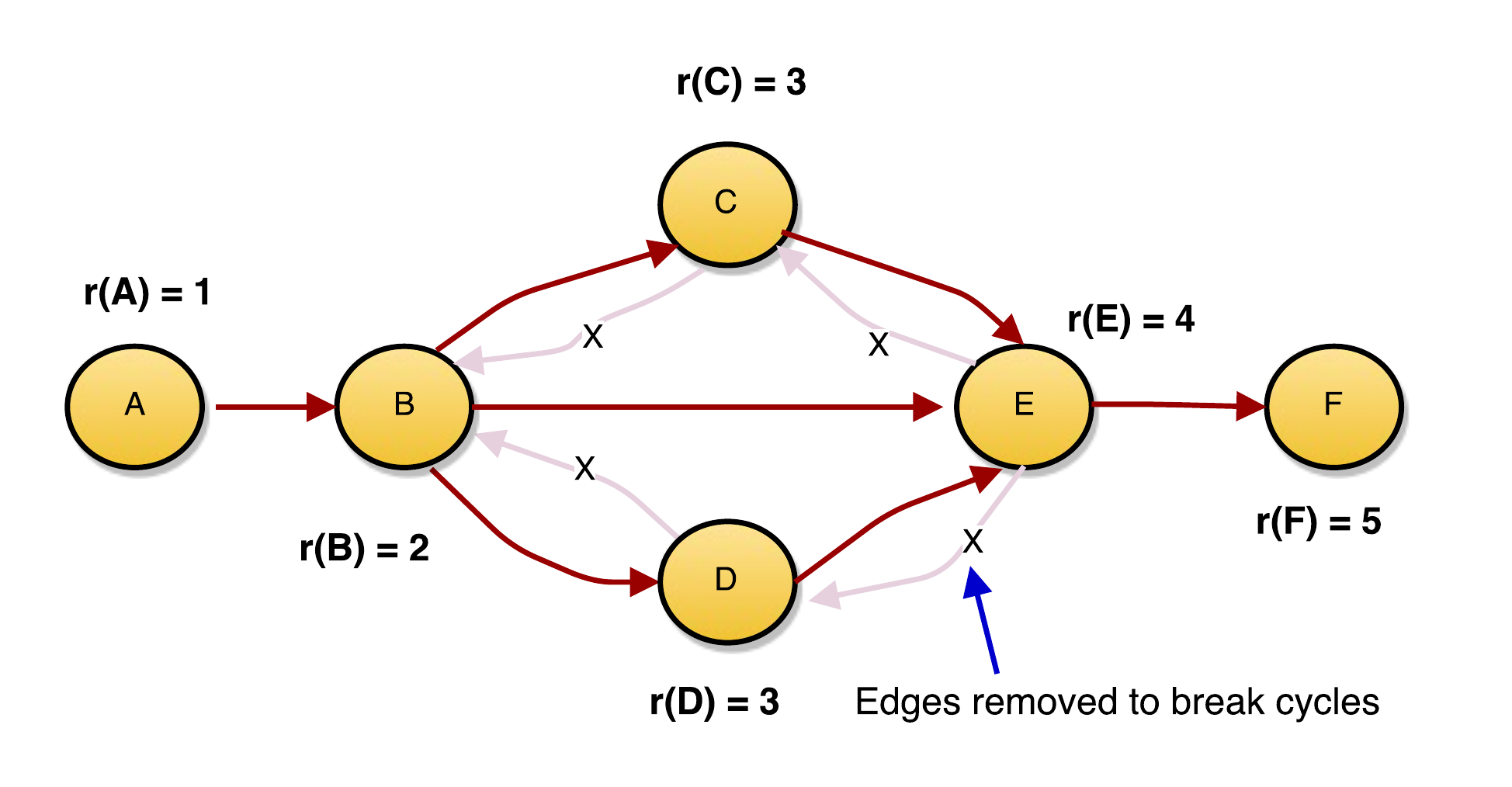}%\vspace{-0.08in}
   \caption{\small Illustration of the limitation of using high order proximities for preserving asymmetric transitivity due to the existence of cycles in the graph: the KI proximity from $B$ to $E$ (inner product between the source vector $u_B^s$ and the target vector $u_E^t$, generated by HOPE) represented as $KI(B,E)$ is $0.0041$, which is smaller than the KI proximity from $E$ to $B$ (inner product between the source vector $u_E^s$ and the target vector $u_B^t$) represented as $KI(E,B) = 0.0067$. HOPE will predict the edge direction is from $E$ to $B$, which is opposite to the real edge $(B,E)$. A similar problem occurs with RPR too, since $RPR(B,E)=0.2129$, which is smaller than $RPR(E,B) = 0.2446$. Due to the existence of cycles among node $B$, $C$, $D$, and $E$, CN and AA are the same for node pair $(B,E)$ and $(E,B)$. The AA proximity predicted by HOPE are $AA(B,E) = AA(E,B) = 0.5$, and CN proximity are $CN(B,E) = CN(E,B) =0$. Hence neither AA or CN can make a confident prediction for the transitivity between $B$ and $E$. However, with using our framework ATP, we can address above limitation. For example, ATP can predict $ATP(B,E) = 1.48$ and $ATP(E,B) = 8.18e^{-10}$, which strongly indicates that the edge direction is from $B$ to $E$.}
    \label{fig:inferring_graph_hierarchy}
%   \vspace{-0.15in}
\end{figure}

HOPE ~\cite{Ou2016KDDAsymmetric}, one of the state-of-art directed graph embedding methods, relies on high-order proximity features (e.g. Adamic Adar (AA), Katz Index (KI), Common Neighbors (CN)) to approximate asymmetric transitivity. %Adamic Adar (AA)~\cite{AdaicAdar}, Common Neighbors (CN), Rooted PageRank (RPR) \cite{Haveliwala2002} and Katz index (KI)~\cite{Katz1953}. %For a directed graph and a pair of vertices $(v_i,v_j)$, Common Neighbors (CN) counts the number of vertexes which is the target of an edge from $v_i$ and the source of an edge to $v_j$. Adamic-Adar (AA) is a variant of common neighbors. Unlike CN, AA assigns each neighbor a weight, which is the reciprocal of the degree of the neighbor. It means that the more vertexes one vertex connected to, the less important it's on evaluating the proximity between a pair of vertex. RPR represents the probability that a random walk from node $v_i$ will locate at $v_j$ in the steady state. KI is a weighted summation over the path set between $v_i$ and $v_j$. The weight of a path is a exponential function of its length. 
%Zhou et al. ~\cite{Zhou2017ScalableGE} proposed a graph embedding method named as APP via random walk with restart, which takes the sampled path as a directed sequence and only observes positive vertex pair along the forward direction. 
%Theoretical analysis of APP shows that the Rooted PageRank (RPR), another higher-order proximity feature, is implicitly preserved in such an embedding.
Zhou et al. ~\cite{Zhou2017ScalableGE} proposed a random walk based graph embedding method named as APP which can implicitly preserve the Rooted PageRank (RPR), another higher-order proximity feature,  in the embedding space.
However, cycles in directed graphs are very common, and these cycles can undermine the performance of embedding strategies such as HOPE and APP, and hence severely limit the capability of the learned embedding vectors in graph inference and analysis. Figure~\ref{fig:inferring_graph_hierarchy} illustrates the limitation of using high order proximities for preserving asymmetric transitivity. %Take Figure~\ref{fig:inferring_graph_hierarchy} for an example\footnote{Edges $(C,B)$, $(D,B)$, $(E,C)$, and $(E,D)$ are in the graph}, the KI proximity from $B$ to $E$ (inner product between the source vector $u_B^s$ and the target vector $u_E^t$, embedding vectors generated by HOPE) represented as $KI(B,E)$ is $0.004153$, which is smaller than the KI proximity from $E$ to $B$ (inner product between the source vector $u_E^s$ and the target vector $u_B^t$) represented as $KI(E,B) = 0.006727$. HOPE will predict the edge direction is from $E$ to $B$, which is opposite to the real direction, since edge $(B,E)$ is in the graph. RPR proximity will make a wrong prediction as KI proximity, since $RPR(B,E)=0.212919$, which is smaller than $RPR(E,B) = 0.244630$. Due to the existence of cycles among node $B$, $C$, $D$, and $E$, CN and AA are the same for node pair $(B,E)$ and $(E,B)$. The AA proximity predicted by HOPE are $AA(B,E) = AA(E,B) = 0.5$, and CN proximity are $CN(B,E) = CN(E,B) =0$. Hence neither AA proximity or CN proximity can make a confident prediction for the edge direction between $B$ and $E$.

% \begin{figure}[ht]
%     \centering
%     \small
%     \includegraphics[width=0.45\textwidth]{pics/HOPE_Drawback_Illustration}
%   \caption{\small Illustration of the limitation for using high order proximities for preserving asymmetric transitivity. HOPE proposed that CN, AA, KI and RPR are highly correlated with asymmetric transitivity. However it has challenges for graphs with cycles. For example, the KI proximity from $B$ to $E$ (inner product between the source vector $u_B^s$ and the target vector $u_E^t$, embedding vectors generated by HOPE) represented as $KI(B,E)$ is $0.004153$, which is smaller than the KI proximity from $E$ to $B$ (inner product between the source vector $u_E^s$ and the target vector $u_B^t$) represented as $KI(E,B) = 0.006727$. HOPE will predict the edge direction is from $E$ to $B$, which is opposite to the real direction, since edge $(B,E)$ is in the graph. RPR proximity will make a wrong prediction as KI proximity, since $RPR(B,E)=0.212919$, which is smaller than $RPR(E,B) = 0.244630$. Due to the existence of cycles among node $B$, $C$, $D$, and $E$, CN and AA are the same for node pair $(B,E)$ and $(E,B)$. The AA proximity predicted by HOPE are $AA(B,E) = AA(E,B) = 0.5$, and CN proximity are $CN(B,E) = CN(E,B) =0$. Hence neither AA proximity or CN proximity can make a confident prediction for the edge direction between $B$ and $E$.} 
%     \label{fig:hope_drawback}
%     \vspace{-0.15in}
% \end{figure}

A strong hierarchical structure in the context of directed networks can help explain complex interactions in many real-world phenomena~\cite{tatti2015}, including asymmetric transitivity. Each node can be assigned a ranking score to represent where it stands in the entire network. The relationship among nodes in such a scenario is fully transitive. For example, if a node $i$ has a lower hierarchy than $j$, and $j$ has a lower hierarchy than $k$, we then can infer that $i$ must have a lower hierarchy than $k$. In graphs with strong hierarchical structure, edges are expected to flow from lower hierarchies to higher hierarchies ~\cite{Gupte2011agony,tatti2015}. However, when transitivity is being predicted leveraging the graph hierarchy alone, without incorporating the inherent graph reachability property, it can sometimes lead to false positive predictions. For example, a lower hierarchy node in a subgraph may not reach a higher hierarchy node in another subgraph which has no connection with the previous subgraph. To redress such problems, one may want to explicitly account for graph reachability as discussed next.

Transitive closure (TC) of a directed graph is a methodology (usually housed in a simple data structure) that makes it possible to answer reachability questions. The TC of a graph $G = (V,E)$ is a graph $G^{+} = (V,E^{+})$ such that for all $v$, $w$ in $V$ there is an edge $(v,w)$ in $E^{+}$ if and only if there is a non-null path from $v$ to $w$ in $G$. However, computing TC for large directed graphs with cycles is expensive, while computing TC of directed acyclic graphs (DAGs) is practical ~\cite{SIMON1988325}. To leverage the above intuition, we propose to first remove a subset of cycle edges which violate the graph hierarchy to reduce a directed graph to a DAG and then leverage the TC of the reduced DAG to represent graph reachability. %One possible way to reduce $G$ to a DAG is to contract each strongly connected component (SCC) in G into a single node. The resulted graph is the condensation of $G$, which provides a simplified view of the connectivity between components. However, nodes in the same SCC share the same hierarchy, which makes it impossible to predict transitivity between nodes in the same SCC. 
We examined several strategies for breaking cycles while preserving the graph hierarchy as much as possible~\cite{trueskill2007,tatti2015}, and found an ensemble approach proposed by Jiankai et al. ~\cite{Sun2017} coupling some of these approaches can meet our requirements. Another benefit of breaking cycles is that the reduced DAG has a very strict hierarchy, and each vertex can be assigned a ranking score effectively and efficiently. %, inferred by a range of features, including a Bayesian skill rating system \cite{trueskill2007} and a social agony metric \cite{Gupte2011agony}, as much as possible. 

%In this paper, we aim to tackle the problem of directed graph embedding with asymmetric transitivity preservation and then leverage the proposed embedding method to solve some fundamental problems in crowdsourcing. 

A key challenge now is how to incorporate graph hierarchy and reachability in a unified framework to preserve the asymmetric transitivity in the embedding space. To this end, we build an asymmetric matrix $\boldsymbol{M}$, which is a non-linear transformation of a diagonal matrix $\boldsymbol{D}$ and an adjacency matrix $\boldsymbol{A}$. Here, $\boldsymbol{A}$ is the adjacency matrix of the transitive closure of the reduced DAG which implicitly contains graph reachability information, and $\boldsymbol{D}$ is a diagonal matrix storing the nodes' hierarchical ranking score along the diagonal entries of a square matrix. Then a factorization based method is applied to $\boldsymbol{M}$ to generate approximate embeddings. In our experiments, an efficient non-negative matrix factorization (NMF)~\cite{Cheng2017LRA} using Cyclic Coordinate Descent(CCD) ~\cite{Nisa2017} with appropriate regularization is leveraged to generate the embedding. Two embedding vectors, source and target vector, are learned for each node to capture the asymmetric transitivity. Through the time complexity analysis of all procedures in our proposed {\bf A}symmetric {\bf T}ransitivity {\bf P}reserving graph embedding framework ({\bf ATP}), we demonstrate that ATP can be applied to large directed graphs efficiently. 

We also conducted extensive experiments to verify the usefulness and generality of the learned embedding in various tasks such as link prediction, and question difficulty estimation and expert finding in online CQAs such as Stack Exchange sites. Particularly, ATP can support inductive embedding learning for newly posted questions (unseen nodes during training), and therefore can route and assign these kinds of questions to approximate experts in CQAs, which tackles a fundamental challenge in crowdsourcing. %Our proposed method has great potential to increase question answering quality, satisfaction, helping all the users to improve their expertise (and not just top-level experts), and reduce inefficiency in real-time crowdsourcing of answers. % Moreover, we believe that our results provide a useful guidance for future designs of collaborative question answer systems, with respect to routing questions to users with matching expertise and interest.
\looseness=-1

\section{Related Works}
\label{sec:relatedWork}

Graph embedding approaches fall into three broad categories classified by Goyal et al. \cite{goyal2017graph}: (1) Factorization based, (2) Random Walk based \cite{kdd14deepwalk,TADW,Gao2018BBN}, and (3) Deep Learning based \cite{Pan2016,Dong2017MSR,SEANO}. Our proposed ATP is factorization based,
and hence we focus on discussing about factorization based techniques in this section. 

Factorization based graph embedding usually solves the graph embedding problem in two steps as follows: (1) represent the connections between nodes in the form of a matrix, and (2) factorize the matrix to get a set of node embedding~\cite{Cai2017Graph,goyal2017graph}. Based on how we construct the input matrix, matrix factorization based approaches are categorized into two types: One is to factorize graph Laplacian, and the other is to directly factorize the node proximity matrix \cite{Cai2017Graph}. The node proximity is preserved by minimizing the loss during the factorizing the node proximity matrix. 

It has been recently shown that many popular random walk based approaches such as DeepWalk ~\cite{kdd14deepwalk}, LINE ~\cite{Tang2015Line}, and node2vec ~\cite{grovernode2vec} can be unified into the matrix factorization framework with closed forms ~\cite{Tang2017EMasMF}. %All of the aforementioned models are based on the Skip-Gram model introduced by Mikolov et al.~\cite{Mikolov2013}. % For example, DeepWalk samples multiple paths from the graph, each of which is regarded as a word sequence. Node2vec offers a flexible sampling strategy, with two parameters controlling the shape of the sampled paths. For each vertex in the sequence, they predict the nearby vertices in both direction, and update the vector according to the Skip-Gram model. 
However, these methods ignore the asymmetric nature of the path sampling procedure and train the model symmetrically, which restricts their applications. Since node pairs from two hop away will be regarded as negative labels, LINE can only preserve symmetric second-order proximity when applied to directed graphs~\cite{Zhou2017ScalableGE}.

%Line introduces the second order proximity between a pair of vertices, which encodes the similarity measured by their local neighborhood. However, Line can only preserve symmetric second-order proximity when applied to directed graphs \cite{Ou2016KDDAsymmetric}. In addition, it cannot preserve the higher-order similarities, since node pairs from two hop away will be regarded as negative labels \cite{Zhou2017ScalableGE}. 

Higher order proximity is considered by many traditional similarity measurements, and has been shown to be effective in many real world tasks. HOPE ~\cite{Ou2016KDDAsymmetric} proposed to use high-order proximities (AA, CN, RPR, and KI) to approximate asymmetric transitivity. % The high order proximities HOPE used are Adamic Adar (AA), Common Neighbors (CN), Rooted PageRank (RPR) \cite{Haveliwala2002} and Katz index (KI). % For a directed graph and a pair of vertices $(v_i,v_j)$, Common Neighbors (CN) counts the number of vertexes which is the target of an edge from $v_i$ and the source of an edge to $v_j$. Adamic-Adar (AA) is a variant of common neighbors. Unlike CN, AA assigns each neighbor a weight, which is the reciprocal of the degree of the neighbor. It means that the more vertexes one vertex connected to, the less important it's on evaluating the proximity between a pair of vertices. RPR represents the probability that a random walk from node $v_i$ will locate at $v_j$ in the steady state. KI is a weighted summation over the path set between $v_i$ and $v_j$. The weight of a path is an exponential function of its length. 
%Zhou et al. \cite{Zhou2017ScalableGE} proposed a graph embedding method APP via random walk with restart, which takes the sampled path as a directed sequence which only observes positive vertex pair along the forward direction.
Theoretical analysis shows that APP implicitly preserves the RPR~\cite{Zhou2017ScalableGE}. %These high order proximities are defined to be asymmetric in directed graphs and assumed to represent the asymmetric structural relationships between vertex pairs.  
However cycles in directed graphs as shown in Figure~\ref{fig:inferring_graph_hierarchy} can hurt the performance of asymmetric transitivity preserving for HOPE and APP, and hence severely limit the capability of the learned embedding vectors in graph inference and analysis.

% All acronyms and their corresponding meanings are shown in Table~\ref{tab:acronyms}.
% \input{tables/acronyms.tex}

%The asymmetric proximity is preserved by factorization based approach is that each node can have two different roles after factorization, the source role and the target role, represented by vector $\vec{s}_v$ and $\vec{t}_v$ respectively. The proximity of vertex pair $(u,v)$ is represented as the inner product of $\vec{s}_u$ and $\vec{t}_v$.

% \input{tex/methodology}

%\vspace{-0.135in}
\section{Our Framework ATP}
\label{sec:methodolody}

\begin{figure*}[!ht]
    \centering
    \small
    \includegraphics[width=0.95\textwidth]{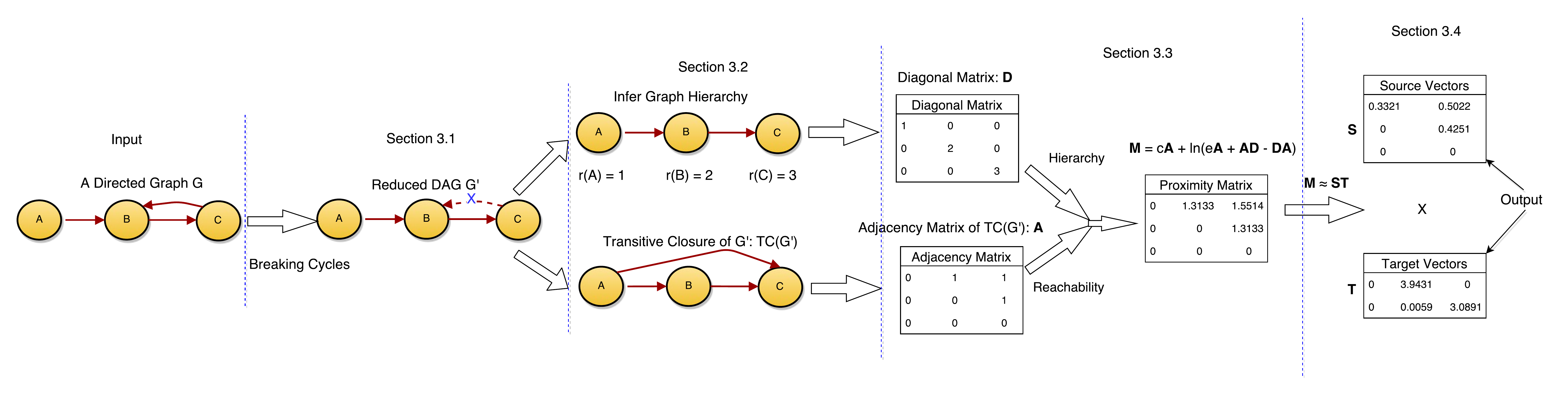}
   \caption{\small Illustration of {\bf A}symmetric {\bf T}ransitivity {\bf P}reserving ({\bf ATP}) graph embedding framework} 
    \label{fig:atp_framework}
    % \vspace{-0.185in}
\end{figure*}

We now describe the \textbf{4-step work-flow of ATP framework}% (Section~\ref{sec:breaking_cycles} -~\ref{sec:mf})
, which is illustrated in Figure-\ref{fig:atp_framework} with a toy example, and then present the computational complexity of our methodology. %in Section~\ref{sec:complexityAnalysis}.  
In the \textbf{first step}  (Section 3.1: Breaking Cycles), an input directed graph $G$ is reduced to a DAG $G'$ by removing a small set of cycle edges which violate the graph hierarchy. 
Then in the \textbf{second step} (Section 3.2: Inferring Graph Hierarchy), each node is assigned a ranking score efficiently based on the hierarchical structure of $G'$. 
The \textbf{third step} (Section 3.3: Incorporating Hierarchy and Reachability) involves the construction of the proposed novel objective matrix $\boldsymbol{M}$ to incorporate both graph hierarchy and reachability information.
Nodes' hierarchical information can be represented using a diagonal matrix $\boldsymbol{D}$, while the transitive closure of $G'$ is represented by $\boldsymbol{A}$.
These two matrices ($\boldsymbol{A}$ and $\boldsymbol{D}$) are used to build the matrix $\boldsymbol{M}$ using a non-linear transformation which can preserve hierarchical rankings between local nodes much better than an ordinary linear model.
The \textbf{final step} (Section 3.4: Generating Asymmetric Transitivity Preserving Graph Embedding from $\boldsymbol{M}$) involves the efficient application of NMF on $\boldsymbol{M}$ to produce two matrices $\boldsymbol{S}$ and $\boldsymbol{T}$, which can be interpreted as asymmetric transitivity preserving source vectors and target vectors of all the nodes in the graph.\looseness=-1
%We next discuss each of these steps in detail.

% \begin{itemize}[leftmargin=0.4cm,noitemsep]
%     \item Remove cycle edges in graph $G$ ($n$ nodes, $m$ edges) to get a DAG $G'$ by using the technique proposed by Sun et al. \cite{Sun2017}
%     \item Assign ranking scores to each node in $G$ ($G'$) based on the structure of $G'$
%     \item Build a matrix $M \in \mathcal{R}^{n \times n}$, of which each element $M_{i,j}$ represents reachability and hierarchical differences between node $i$ and $j$
%     \item Matrix factorization of $M \approx ST$, where $S \in \mathcal{R}^{n \times k}$ and $T \in \mathcal{R}^{k \times n}$. Each row in  $S$ represents a node's latent feature as a source node, and each column in $T$ represents a node's latent feature as a target node
% \end{itemize}

\subsection{Breaking Cycles}
\label{sec:breaking_cycles}

Reducing a directed graph $G = (V,E)$ to a DAG $G' = (V,E')$ has two obvious advantages: 1) making it possible for us to compute the transitive closure of $G'$; 2) inferring the hierarchy of $G'$ becomes easier, since a DAG has a very strict hierarchy. %In this subsection, we discuss possible strategies to break cycles.
\looseness=-1

We examined several strategies for breaking cycles while preserving the graph hierarchy as much as possible, and found an ensemble approach {\em H\_Voting}  proposed by Jiankai  et al. ~\cite{Sun2017} can meet our requirements. {\em H\_Voting} selects the edge with the highest voting score for removal in a fast, scalable, and fully automated way. The voting score of each edge is determined by the severity of their violation, which means that edges that respect the hierarchy receive a score of $0$ and score increase linearly as the hierarchy violation becomes more severe. The corresponding hierarchy is inferred by ensembling TrueSkill~\cite{trueskill2007} and Social Agony~\cite{Gupte2011agony,tatti2015}. Figure~\ref{fig:inferring_graph_hierarchy} illustrates that edges (C,B), (D,B), (E,C) and (E,D) are removed  by {\em H\_Voting} to break cycles. %We also examined two other voting strategies - {\em SA\_Voting} and {\em TS\_Voting}, which ensembles $3$ Social Agony and $3$ TrueSkill based strategies %({\em SA\_G}, {\em SA\_B}, {\em SA\_F}) and ({\em TS\_G}, {\em TS\_B}, {\em TS\_F}) 
%respectively in our experiments. 
Empirically, it has been shown that {\em H\_Voting}, can accurately identify the edges to be removed, even in noisy and large-scale real-world graphs. The time complexity of breaking cycles is $O(E^2)$ in the worst case, which happens in directed complete graphs \footnote{Every pair of distinct vertices is connected by a pair of unique edges (one in each direction).}.\looseness=-1

\subsection{Inferring Graph Hierarchy}
\label{sec:inferring_graph_hierarchy}

Given that the graph has been converted to a DAG using the previous step, graph hierarchy can be inferred based on this reduced DAG. Given a graph $G = (V,E)$, inferring graph hierarchy means that we have to construct a function $r: V \rightarrow \mathbb{Z}$, which maps each vertex to an integer, representing corresponding vertex's hierarchy in $G$. The computed graph hierarchy is fully transitive and can be used to infer the asymmetric transitivity in $G$.
%The state-of-the-art method to infer graph hierarchy is by leveraging Social Agony, which's proposed by Gupte et al.~\cite{Gupte2011agony} assuming that agony can be caused when directed edges go from higher ranking nodes to lower ranking nodes. Then the the problem of inferring graph hierarchy is changed to find a ranking $r$ that minimize the total agony in the graph. Gupte et al. provided an $O(VE^2)$ algorithm to minimize the agony of the graph~\cite{Gupte2011agony}. A faster discovery algorithm with the computational complexity of $O(E^2)$ was introduced by Tatti~\cite{Tatti2014,tatti2015}.  
%To reduce the time complexity of inferring graph hierarchy, we can leverage a nice property of DAGs that they have strict hierarchies.
%A nice property of DAGs is that they have strict hierarchical structures, which can be leveraged to infer their hierarchy. 
One straightforward way to compute a ranking score for each vertex is to use topological sorting. % with time complexity $O(V+E)$. 
However, topological sorting is non-deterministic. Hence we modify topological sorting algorithm by assigning a ranking score to each vertex in a DAG recursively following the steps below: Step 1) assign the current ranking score $o$ \footnote{Ranking score $o$ is initialized to $1$.} to all nodes with zero in-degree; Step 2) update the target graph by removing all zero in-degree nodes and their corresponding out-going edges; Step 3) update the current ranking score $o$ by increasing $1$.

% \begin{itemize}[leftmargin=0.4cm,noitemsep]
%     \item Step 1: assign the current ranking score $o$ \footnote{Ranking score $o$ is initialized to $1$.} to all nodes with zero in-degree.
%     \item Step 2: update the target graph by removing all zero in-degree nodes and their corresponding out-going edges.
%     \item Step 3: update the current ranking score $o$ by increasing $1$.
% \end{itemize}

Figure~\ref{fig:inferring_graph_hierarchy} shows each node's ranking score inferred by the above procedures. For example, node $A$'s in-degree in the reduced DAG is $0$ while its hierarchy is represented as $r(A) = 1$. The time complexity of above procedure is the same as topological sorting, which is $O(|V|+|E|)$.\looseness=-1

% We use Algorithm~\ref{alg:ranking_orders} to assign ranking scores to each vertex in $G$ ($G'$).

% \begin{algorithm}[h]
% \small
% \SetAlgoLined
% \KwResult{Each node will be assigned a ranking score based on its hierarchy in the DAG $G'$}

%  order = 1\;
 
%  \While{$G' \neq \emptyset$}{
%  delete\_nodes = \emptyset\;
 
%   \For{each node $v \in G'$ and indegree(v) = 0} {
%   r(v) = order\;
%   order += 1\;
%   delete\_nodes.add(v)\;
%   }\
%   $G'$ -= delete\_nodes\;
%   }
% \caption{Assign hierarchical ranking scores to vertices}
% \label{alg:ranking_orders}
% \end{algorithm}

% In Algorithm~\ref{alg:ranking_orders}, $indegree(v)$ represents the indegree of node $v$ in DAG $G'$. And if node $u$ can reach $v$ in $G'$, then we can infer that $r(u) < r(v)$. The time complexity of this procedure is same as topological sort, which is $O(V+E)$. Figure~\ref{fig:inferring_graph_hierarchy} illustrates how we remove cycle edges and assign ranking scores to nodes in the graph.

% \input{tex/ourApproach/buildMatrix}

\subsection{Incorporating Hierarchy and Reachability}
%\subsection{Building Hierarchical Proximity Matrix $\boldsymbol{M}$}
\label{sec:build_matrix}

The hierarchical ranking score inferred by the methodology in Section 3.2 reflects where a node stands in the entire network, and it is fully transitive. However, the sole assumption that edges are from lower to higher hierarchy nodes fails to incorporate the inherent graph reachability property, and hence is prone to generating many false positive predictions during down-stream analysis tasks (eg: link prediction using the generated node embeddings).
Thus the key challenge is to combine both graph hierarchy and reachability inside a unified framework, which we discuss next.\looseness=-1

%In Section~\ref{sec:inferring_graph_hierarchy}, we have shown how to infer the graph hierarchy based on the reduced DAG. In this section, we show how to leverage graph reachability efficiently.\looseness=-1

TC can be thought of as constructing a data structure that makes it possible to answer reachability questions. Instead of computing TC of the original directed graph $G$, we compute the TC of the reduced DAG $G'$ and represent it as its adjacency matrix $\boldsymbol{A} \in \mathbb{R}^{|V| \times |V|}$. If node $i$ can reach $j$ in $G'$, then the corresponding element $A_{i,j} = 1$, otherwise $A_{i,j} = 0$.
% \begin{equation}
% A_{i,j} = \begin{cases} $1$, & \mbox{if $i$ can reach  $j$  in  DAG $G'$} \\ 
% 0, & \mbox{otherwise} \end{cases}
% \label{eq:constant}
% \end{equation}
%$\boldsymbol{A}$ is the adjacency matrix representation of the transitive closure of $G'$. 
% Simon proposed an efficient algorithm for computing transitive closure of DAGs which runs in worst-case time $O(k \cdot e_{red})$ and space $O(V \cdot k)$, where $e_{red}$ is the number of edges in the transitive reduction of $G'$, and $k$ is the width of a chain decomposition  \cite{SIMON1988325}.
% For the expected values in a $G_{V,p}$ model of a random DAG with $0 < p < 1$, we have $E(k) = \frac{log(p \cdot V)}{V}$, $E(e_{red}) = O(V \cdot log(V))$, and if $log^2(V)/V \leq p$, $E(k \cdot e_{red}) =  O(V^2)$, otherwise $O(V^2loglog(V))$.
% \begin{equation}
% E(k \cdot e_{red}) = \begin{cases} O(V^2), & \mbox{$log^2(V)/V \leq p$} \\ 
% O(V^2loglog(V)), & \mbox{otherwise} \end{cases}
% \label{eq:time_complexity}
% \end{equation}
$\boldsymbol{A}$ contains the information of graph reachability, but it treats the hierarchical difference between any reachable node pairs the same (equal to $1$). To emphasize the impact of these non-zero elements in $\boldsymbol{A}$ and leverage graph hierarchy, a simple way is to replace each non-zero element by corresponding node pair's hierarchical difference, which is equivalent to applying a linear function to transform $\boldsymbol{A}$ to $\boldsymbol{L} \in \mathbb{R}^{|V| \times |V|}$. %, which can measure the ranking differences between reachable node pairs in $G'$. 
For each non-zero element $A_{i,j} = 1$ in $\boldsymbol{A}$, its corresponding $L_{i,j}$ in $\boldsymbol{L}$ is $\Delta_{i,j} = r(j) - r(i)$ and $\Delta_{i,j} \geq 1$.  \looseness=-1 
% \begin{equation}
% L_{i,j} = \begin{cases} r(j) - r(i), & \mbox{if $i$ can reach  $j$  in  DAG $G'$} \\ 
% 0, & \mbox{otherwise} \end{cases}
% \label{eq:linear}
% \end{equation}
% Where every $L_{i,j} >= 1$.

Suppose $\boldsymbol{D} \in \mathbb{R}^{|V| \times |V|}$ is a diagonal matrix, where each non-zero element in the diagonal is $D_{i,i} = r(i)$. Then we have:
% \begin{equation}
% R_{i,j} = \begin{cases} r(i), & \mbox{if $i = j$} \\ 
% 0, & \mbox{otherwise} \end{cases}
% \label{eq:diagonal}
% \end{equation}

% \vspace{-0.05in}
\begin{equation}
    \boldsymbol{L} = \boldsymbol{AD} - \boldsymbol{DA}
    \label{eq:linear}
\end{equation}

%A small value of $\Delta$ in $\boldsymbol{L}$ represents a local view of nodes' relationship in the graph, while a relatively large value of $\Delta$ shows a global view of the graph's structure.
Empirically, the maximum value of $\Delta$ in $\boldsymbol{L}$ is much larger than the minimum value (which is $1$). %For example, in our experiments, the maximum value of $\Delta$ in Cit-HepPh, and GNU graph are $187$, and $267$ respectively. 
The high (and varying) range of values in $\Delta$ will unfavourably amplify the effect of large hierarchical difference values while damping the effects of smaller values, thereby negatively impacts the transitivity preserving property in local sub-graphs.
To overcome this limitation, we seek to reduce the absolute values of $\Delta$ to smaller ones, while preserving its important monotonic property.
We observe that a simple yet popular harmonic series, which is a non-linear and non-decreasing function, can satisfy our requirements very well and can be used to build the proximity matrix $\boldsymbol{M} \in \mathbb{R}^{|V| \times |V|}$.
Each non-zero entry $L_{i,j} \in \boldsymbol{L}$ is transformed to $M_{i,j} =  1+\frac{1}{2}+...+\frac{1}{\Delta_{i,j}}$ in $\boldsymbol{M}$. Since a harmonic number $h(\Delta_{i,j}) = \sum_{k=1}^{\Delta_{i,j}} \frac{1}{k}$ can be approximated by $(\gamma + log(\Delta_{i,j}))$~\footnote{$\gamma$ is the Euler-Mascheroni constant, log is the Natural logarithm}, without loss of generality, we represent each non-zero element $M_{i,j} = c + log(e + \Delta_{i,j})$, where $c$ is a constant, and $e$ is the  mathematical constant satisfying $log(e + \Delta) > 0$.
Figure~\ref{fig:atp_framework} provides an example of computing $\boldsymbol{M}$, where $c = 0$.
After this non-linear transformation, the gap of hierarchical rankings between local nodes can be noticed and well preserved compared to the linear model. % \textbf{(WHY? Provide some intuition or example.)}
Thus, the final matrix $\boldsymbol{M}$ incorporating both graph hierarchy and reachability, 
%which will be used for generating embedding,
is computed as:\looseness=-1

\begin{equation}
    \boldsymbol{M} =  c \boldsymbol{A} + log(e\boldsymbol{A} + \boldsymbol{L}) = c\boldsymbol{A} + log(e\boldsymbol{A} + \boldsymbol{AD} -\boldsymbol{DA})
\end{equation}\looseness=-1

Constructing the adjacency matrix $\boldsymbol{A}$ is equivalent to computing the transitive closure of $G'$, which has a worst-case time complexity of $O(|V|^2loglog(|V|))$~\cite{SIMON1988325}.
%Constructing the adjacency matrix $\boldsymbol{A}$ is equivalent to computing the transitive closure of $G'$, which has a worst-case time complexity of $O(k e_{red})$ and space complexity of $O(V k)$ \footnote{ $e_{red}$ is the number of edges in the transitive reduction of $G'$, and $k$ is the width of a chain decomposition}\cite{SIMON1988325}.  For the expected values in a $G_{V,p}$ model of a random DAG with $0 < p < 1$, we have $E(k) = \frac{log(p V)}{V}$, and $E(e_{red}) = O(Vlog(V))$. If $log^2(V)/V \leq p$, $E(k  e_{red}) =  O(V^2)$, otherwise $O(V^2loglog(V))$.
%The time complexity of the current fastest matrix multiplication algorithm is $O(V^{2.38})$ \cite{Coppersmith1987}. 
The time complexity of computing  $\boldsymbol{L}$ with Equation \ref{eq:linear} is equal to the number of non-zero elements in $\boldsymbol{A}$, which is $O(|V|^2)$ in the worst case. Hence, the time complexity of constructing $\boldsymbol{M}$ is $O(|V|^2loglog(|V|))$ in the worst case. %\looseness=-1 

\subsection{Generating Asymmetric Transitivity Preserving Graph Embedding from $\boldsymbol{M}$}
\label{sec:mf}
We have discussed how to build the non-negative proximity matrix $\boldsymbol{M}$, which is a function of the adjacency matrix of the transitive closure of the reduced DAG and a diagonal matrix which contains graph hierarchy. In this section, we propose to use factorization models to generate asymmetric transitivity preserving embedding for the given directed graph. %Some state-of-art factorization models such as Non-negative Matrix Factorization (NMF) and Factorization Machine (FM) can be leveraged naturally to generate the desired embeddings.

% \subsubsection{Generating Graph Embedding via NMF}
% \label{sec:atp_nmf}

The most straightforward way is to apply NMF on $\boldsymbol{M}$ to generate asymmetric transitivity preserving embedding for the given directed graph.
NMF of the $(|V| \times |V|)$ matrix $\boldsymbol{M}$ generates a low-rank approximation of it: $\boldsymbol{M} \approx \boldsymbol{ST}$, where $\boldsymbol{S} \in \mathbb{R}^{|V| \times k}$ and $\boldsymbol{T} \in \mathbb{R}^{k \times |V|}$, as shown in Figure~\ref{fig:atp_framework}.
Each row in $\boldsymbol{S}$ represents a node's out-reach (source) vector, and each column in $\boldsymbol{T}$ represents a node's in-reach (target) vector. $k$ is the dimension size of the source/target embedding space. NMF step is a core part in generating the embedding and we apply a multi-core GPU version of NMF using CCD GPUCCD++~\cite{Nisa2017} with appropriate regularization to generate the embedding for large graphs efficiently. The time complexity per iteration of GPUCCD++ is $O(k|V|^2)$  in the worst case. %~\cite{CCD2014}.%\looseness=-1

To predict whether there is a directed path from node $i$ to node $j$, we check the value of $\sigma (\langle \vec{s}_i , \vec{t}_j \rangle)$, where $\sigma$ is the sigmoid function, $\vec{s}_i$ is node $i$'s source vector and $\vec{t}_j$ is node $j$'s target vector respectively. If $\sigma (\langle \vec{s}_i , \vec{t}_j \rangle) > \alpha$, there is a predicted path from $i$ to $j$. $\alpha$ is a threshold with range in $[0.5,1)$. We set $\alpha = 0.5$ in our experiments, which we empirically found to work well.

\subsection{Complexity Analysis}
\label{sec:complexityAnalysis}

In this section, we analyze the complexity of the whole framework of ATP, given a directed graph $G=(V,E)$ as input. Fundamental procedures of ATP are breaking cycles, inferring graph hierarchy, constructing $\boldsymbol{M}$, and factorization of $\boldsymbol{M}$. In the worst case, their corresponding time complexity is $O(|E|^2)$, $O(|E|+|V|)$, $O(|V|^2loglog(|V|))$, and $O(|V|^2k)$ per iteration (NMF) respectively. By combining them, the time complexity of ATP is $O(|E|^2)$ in the worst case ($G$ is a directed complete graph). The bound is very pessimistic in practice, and the bottleneck part (breaking cycles) can be parallelized since it can perform on each SCC independently to remove cycle edges. %Hence, ATP is scalable for large directed graphs. \looseness=-1

\section{Experiments and Analyses}
\label{sec:experiments}

We apply our graph embedding framework ATP to three diverse tasks: link prediction, and  question difficulty estimation and expert finding in CQAs.

\subsection{Link Prediction}
\label{sec:link_prediction}

In link prediction, we would like to predict these missing edges given a network with a certain fraction of edges removed. 
The labeled dataset of edges (or node pairs) consists of positive and negative examples. Given a random edge $e$, if the removal of this edge will not disconnect the residual network, $e$ will be selected as a positive example. We select $r = 10\%$ edges as positive examples. To generate negative examples, we randomly select an equal number of node pairs from the network which have no edges connecting them \footnote{Each node pair $(u,v)$ in negative samples satisfies the condition that $v$ can reach $u$, but $u$ cannot reach $v$ in the network.}. Hence $2r$ edges and node pairs are selected for evaluation.

%Statistics of the datasets we use are shown in Table ~\ref{tab:dataset_lp}. 

Datasets used for evaluation are Wiki-Vote\footnote{\url{https://snap.stanford.edu/data/wiki-Vote.html}}, GNU\footnote{\url{https://snap.stanford.edu/data/p2p-Gnutella31.html}},  Cit-HepPH\footnote{\url{https://snap.stanford.edu/data/cit-HepPh.html}}, which were used in prior work \cite{Lai2017Prune}. GNM-30K is a random directed graph generated with cycles ($30K$ nodes and $155K$ edges).

Following existing literature \cite{grovernode2vec,ijcai2017-fastNetworkEmbedding,Tran2018}, we use Area Under Curve (AUC) %~\cite{Roc1982}
to evaluate the link prediction performance. We compare our method with the most recent work for asymmetric proximity preserving.

\begin{itemize}[leftmargin=0.4cm,noitemsep]
    \item \textbf{ATP and its variants}: ATP-Constant ($\boldsymbol{M}$ = $\boldsymbol{A}$), ATP-Linear ($\boldsymbol{M}$ = $\boldsymbol{L}$), ATP-Harmonic, and ATP-log (By default ATP refers to ATP-log). ATP-Harmonic, and ATP-log transforms $\boldsymbol{L}$ to $\boldsymbol{M}$ by harmonic and log function respectively.  
    \item \textbf{HOPE}\cite{Ou2016KDDAsymmetric}: 
    %HOPE has shown its improvements over DeepWalk \cite{kdd14deepwalk}, Partial Proximity Embedding (PPE) \cite{Song2009PPE}, Common Neighbors (rank the links by the number of common neighbors), and Adamic Adar (rank the links by Admaic-Adar values) in link prediction. 
    As the time complexity of computation of RPR is too high, we only report performances of HOPE-AA, HOPE-CN, and HOPE-KI here. %\footnote{In fact, Ou et al. tested HOPE-RPG only on small-scaled synthetic data in their corresponding paper}. 
    \item \textbf{SVDM}: SVD-Harmonic and SVD-log use the same way to build $\boldsymbol{M}$ as ATP-Harmonic and ATP-log respectively. However, unlike ATP, SVD-Harmonic and SVD-log performs Singular Value Decomposition (SVD) as used in HOPE on $\boldsymbol{M}$ and selects the largest $k$ singular values and corresponding singular vectors to construct the embedding. 
    % \item \textbf{Shortest Path Length (SPL)}: We also have explored the possibility of using SPL as  high order proximity for measuring the asymmetric transitivity. The assumption is that the shorter path from node $i$ to $j$, the more similar should be $i$'s source vector and $j$'s target vector. However, computing all pairs of SPL is expensive for graphs with cycles. Hence we compute the SPL between node pairs based on the reduced DAG $G'$ instead. Like ATP, SPL also leverages graph reachability, since each non-zero element $M_{i,j}$ represents the SPL from node $i$ to $j$ in $G'$. 
    \item \textbf{LINE} \cite{Tang2015Line}: It is worth mentioning that LINE can only preserve symmetric second-order proximity when applied to a directed graph. In our experimental settings, vertex vectors are considered as source vectors, and context vectors are used as target vectors.  
\end{itemize}

\subsubsection{Performance Analysis} 

%In our experiments, we set the embedding dimension $k = 8$ for Wiki-Vote dataset and $k=64$ for all other datasets\footnote{We set a small $k$ for Wiki-Vote, since Wiki-Vote is the smallest graph among three real graphs (GNU, Cit-HepPh, Wiki-Vote). To keep consistent, two random graphs use the same k = 64 as GNU and Cit-HepPh.}. 
We can conclude from the performance as shown in Table~\ref{tab:hge_hope_comparison} that: 

\begin{table}
\centering
\small
%\scriptsize
\caption{Comparisons between ATP and the state-of-the-art methods on link prediction, evaluated by AUC}
\setlength\tabcolsep{2.75pt} % default value: 6pt
 \begin{tabular}{@{}c|c|cccc@{}} \toprule
 & AUC & Wiki-Vote & Cit-HepPH & GNU &  GNM-30K\\ \midrule
LINE & 2-nd order & 0.4423 & 0.3310 & 0.4748 &  0.4109 \\ \midrule
  & AA & 0.7672 & 0.7385 & 0.5565 & 0.5747\\
 HOPE & CN & 0.7860 & 0.7570 & 0.5736 & 0.5836\\
 & AI & 0.7784 & 0.7440 & 0.6159 &  0.5784\\ \midrule 
SVDM & Harmonic  & 0.8200 & 0.7522 & 0.8166  & 0.6255\\ 
    & log       &  0.8215	& 0.7929	& 0.8162	& 0.6248 \\ \midrule 
%Shortest Path Length & G & 0.8419 & NA \footnote{Job is terminated due to time or memory limitation.} & NA & 0.7241 & NA\\
% SPL & DAG  & 0.8847 & 0.8208 & 0.8803 &  0.7422\\ \midrule
 & Constant & 0.9123 & 0.7939 & 0.8684 &  0.7845\\
 & Linear & 0.9462 & 0.8682 & 0.8893 &  0.8530\\
 {\bf ATP} &{\bf log} & {\bf 0.9481} & {\bf 0.8916} & {\bf 0.9314}  & {\bf 0.8789}\\ 
 %&\cellcolor[rgb]{ .851,  .851,  .851}{log} & \cellcolor[rgb]{ .851,  .851,  .851}0.9481 & \cellcolor[rgb]{ .851,  .851,  .851}0.8916 & \cellcolor[rgb]{ .851,  .851,  .851}0.9314 & \cellcolor[rgb]{ .851,  .851,  .851}0.8084 & \cellcolor[rgb]{ .851,  .851,  .851}{0.8789}\\ 
  & Harmonic & 0.9478 & 0.8892 & 0.9288  & 0.8777\\
 \bottomrule
\end{tabular}
\label{tab:hge_hope_comparison}
% \vspace{-0.15in}
\end{table}

\begin{itemize}[leftmargin=0.4cm,noitemsep]
    \item Since Harmonic numbers can be approximated by log functions, log and Harmonic transformation can achieve similar performance in both ATP and SVDM. %The average difference between ATP-log and ATP-Harmonic, and SVD-log and SVD-Harmonic are $0.016\%$ and $1.1\%$ respectively.
    By default, we use log function as our non-linear transformation. It is noticeable that ATP performs better than SVDM. In average, ATP improves over SVDM by $22.01\%$ among all datasets (from $12.45\%$ to $40.67\%$), which shows the advantage of leveraging NMF to perform transitivity preserving graph embedding. 
    
    \item ATP and SVDM perform better than HOPE. In average, SVDM improves over HOPE-KI by $13.72\%$ among all datasets, and ATP improves over HOPE-KI from $19.84\%$ to $51.97\%$. The only difference between SVDM and HOPE is the technique used to build $\boldsymbol{M}$. The results demonstrate the advantage of incorporating graph reachability and hierarchy to construct $\boldsymbol{M}$, in comparison to building a higher order proximity matrix based on AA, CN, and KI.
    
    \item Both ATP-Harmonic and ATP-log perform better than ATP-Linear and ATP-Constant as expected, and ATP-Linear performs better than ATP-Constant. For example, ATP-log improves over ATP-Constant by $8.51\%$ in average among all datasets. ATP-log improves over ATP-Linear by $4.74\%$ on the largest dataset GNU. It shows the efficacy of applying non-linear transformation to $\boldsymbol{M}$.

    % \item ATP performs better than SPL since SPL is not always consistent with the graph transitivity. For example, in a graph with edges $(a,b),(b,c),(c,d),(a,d)$, the SPL between node $a$ and $b$ is $l(a,b) = 1$, and $l(a,c) = 2$. The edge direction is from $b$ to $c$. While $l(a,d) = 1$ and $l(a,c) =2$, the edge direction is from $c$ to $d$. We also tested the performance of using SPL computed from two small networks of Wiki-Vote and GNM-5K, and their corresponding AUC are $0.8419$ and $0.7241$ respectively, which shows that computing all pairs of SPL based on the reduced DAG is not only more efficient but also more effective than doing on the original graph, since breaking cycles can remove some noisy edges~\cite{Sun2017}. It demonstrates the advantage of breaking cycles and indicates that breaking cycles can be used as a generic pre-processing step to other problem domains as well. 
\end{itemize}

\subsection{Question Difficulty Estimation and Expert Finding in CQAs}
\label{sec:question_routing_cqas}

In this section, we start by discussing how to apply ATP to estimate question difficulty and user expertise. We then show how to embed newly posted questions (unseen nodes in the training) inductively and identify best answerers for newly posted questions in CQAs.

\begin{table}[t!]
\small
%\scriptsize
\centering
\caption{Statistics of Stack Exchange Sites} 
\setlength\tabcolsep{1pt} % default value: 6pt
\begin{tabular}{c|cccccccc}
Stack Exchange Sites 　　　&  Apple  & Gaming & Physics & Scifi  & Unix\\ \hline
\# nodes in Graph   　　　& 133K  & 117K & 127K & 59K &   167K\\
\# edges in Graph  　　　 & 161K   & 190K & 188K & 97K &   249K\\
\# questions with bounty  & 1,834  　 & 1,562 & 1,922 　　　& 980 &   1,614 \\
\# cold questions  for evaluation  & 234 & 313 & 196 & 279        & 297 \\ 
\bottomrule
\end{tabular}
\label{tab:statistics_stack_exchange}
% \vspace{-0.15in}
\end{table}

\subsubsection{Question Difficulty and User Expertise Estimation in CQAs}

We first talk about how to apply our graph embedding technique into question difficulty and user expertise estimation, which is a central part of automated question routing in CQAs. We select $5$ large and popular sites from Stack Exchange \footnote{We used the data dump which is released on June 12, 2017 and is available online at \url{https://archive.org/details/stackexchange}} for evaluation. More details about the Stack Exchange sites can be found in the Table~\ref{tab:statistics_stack_exchange}. For each Stack Exchange site, ATP uses the same competition graph as the input as QDEE~\cite{QDEE2018} which assigns solely scalar value to represent question difficulty level and user expertise. In this section, we will show the advantages of learning latent representations for question difficulty (question nodes) and user expertise (user nodes) by leveraging ATP.

 Following the same setting as QDEE~\cite{QDEE2018}, questions which were provided non-zero bounty scores are selected as our ground truth for evaluation, and pairwise accuracy (\textit{Acc})\footnote{$\textrm{Acc =} \frac{\textrm{\# correctly predicted question pairs}}{\textrm{\# all question pairs}}$}, as used by previous studies \cite{wang2014,QDEE2018}, are used to measure the effectiveness of different estimation techniques. %A question pair $(i,j)$ is regarded as correctly predicted if $\sigma (\langle \vec{s}_i , \vec{t}_j \rangle) > \alpha$ and bounty score of $i$ is less than $j$.
 Higher accuracy indicates better performance of the technique. The number of questions with bounty used for evaluation is shown in Table~\ref{tab:statistics_stack_exchange}. We evaluate ATP and other state-of-the-art methods such as {\bf TrueSkill} ~\cite{wang2014}, {\bf Number-Of-Answers}~\cite{yang2014sparrows}, {\bf Time-First-Answer},  {\bf Time-Best-Answer}~\cite{Huna2016} and {\bf QDEE}~\cite{QDEE2018} on the task of question difficulty estimation. %{\em Number-Of-Answers} estimates question difficulty by using the number of answers provided for the target question ~\cite{yang2014sparrows}. {\em Time-First-Answer} ({\em Time-Best-Answer}) estimates question difficulty by measuring how long it takes such that the target question gets its first (best) answer~\cite{Huna2016}.%Hanrahan2012

\begin{figure}[t!]
    \centering
    \small
    \includegraphics[width=0.485\textwidth]{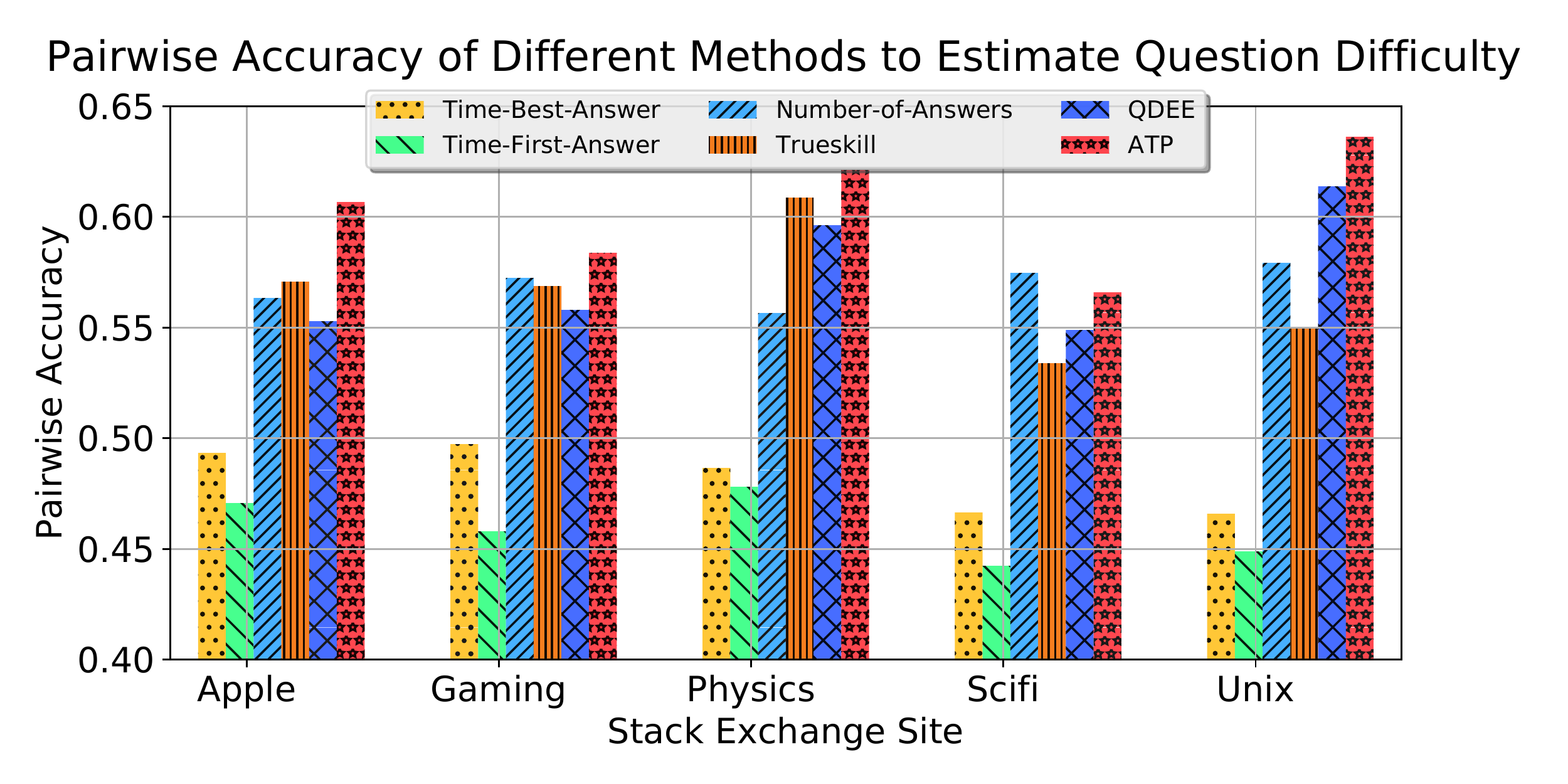}
    \caption{Pairwise accuracy  of different approaches to estimate question difficulty} 
    \label{fig:hge_performance_qde}
%   \vspace{-0.2in}
\end{figure}

Question difficulty estimation performance is shown in Figure~\ref{fig:hge_performance_qde}.
We can conclude that { ATP} performs the best on almost all of the Stack Exchange sites. For example, { ATP} improves over { TrueSkill}, { Number-of-Answer}, and { QDEE} on average by $6.56\%$, $5.92\%$, and $5.06\%$ respectively in $5$ Stack Exchange sites. { TrueSkill} suffers from the data sparsity problem. Each question has only one in-edge (from questioner) and one out-edge (to the best answerer), which limits the accuracy of { TrueSkill}. { ATP} can leverage graph reachability and hence each question can have interactions with other questions and users, which can overcome the data sparseness. %{\em Number-of-Answer} has a comparable performance with {\em ATP} only in site Scifi (\textit{Acc} of {\em Number-of-Answer} and {\em ATP} are $0.5745$, and $0.5659$ respectively) which has the second highest average number of answers per question (ANAPQ) among all $8$ sites. However, a larger value of ANAPQ does not indicate better performance of {\em Number-of-Answer}. For example, {\em ATP} improves over {\em Number-of-Answer} by $11.99\%$ in the Stack Exchange site which has the highest ANAPQ. % which is $12.22$.

\subsubsection{Inductive Embedding Learning for Cold Question Routing in CQAs}

Usually, there are two types of questions in CQAs -- resolved (questions with answers) and newly posted questions (questions that have not received any answers). %The newly posted questions may themselves be posted by new askers (such as new registered users who have not asked a question earlier) or existing askers (such as users who have asked several questions previously).
We refer to these newly posted questions as {\em cold questions}. The majority of approaches have focused on evaluating content quality after the fact (after questions have been resolved) \cite{Yang2013CQArank}. Yet, as the CQAs continue to grow, routing the cold questions to matching experts before answers have been provided has become a critical problem. For example, in Stack Overflow, about 4.8 million questions have not been answered \footnote{\url{https://stackoverflow.com/unanswered}}.

In this section, we show that ATP can generate quality embedding for new vertices (cold questions) unseen during training, therefore supporting inductive learning in nature. Our idea is to leverage Expertise Gain Assumption ({\em EGA}) ~\cite{QDEE2018} to bridge the gap between cold-start and well-resolved questions asked by the same user. Given a cold question $q^*$ asked by a user $u^*$, the most recent $k$ questions asked by the same user $u^*$ are $q_1,q_2,...,q_k$ and their associated embedding are available to us (they can be seen during training). We use the embedding of the question which has the highest difficulty level among $q_1,q_2,...,q_k$ to approximate $q^*$'s embedding. A question $q_{max}$ is considered to have the highest difficulty level if $\sigma (\langle \vec{s}_{q_i} , \vec{t}_{q_{max}} \rangle) > \sigma (\langle \vec{s}_{q_{max}},\vec{t}_{q_{i}} \rangle)$ for all $i \in [1,k]$ and $q_i \neq q_{max}$. Then $\vec{s}_{q^*} = \vec{s}_{q_{max}}$ and  $\vec{t}_{q^*} = \vec{t}_{q_{max}}$. We note that it is possible that the user posing the question is a new user (or one that has not posted a sufficient number of questions). In this case, $k$ well-resolved questions that are closest (i.e. cosine similarity) to $q^*$ in textual descriptions \footnote{Each question can be represented as a feature vector by LDA~\cite{Ji2012LDA}}, are picked as its nearest neighbors. The source and target embedding of $q^*$ is predicted as the averaged source and target embedding of its nearest neighbors respectively. 

Our task of cold question routing is to select the user who has the highest possibility to be selected as the best answerer for a newly posted question. Given the testing question set $\mathcal{Q}_t$, the predicted ranking list of all potential answerers for a test question $q^*$ is $R^{q^*}$ for all $q^* \in \mathcal{Q}_t$. The ranking score of a potential answerer $u$ for the cold question $q^*$ is computed as $\sigma (\langle \vec{s}_{q^*} , \vec{t}_u \rangle)$. The answerer who has the highest ranking score will be selected as the best answerer for $q^*$. %$R^{q^*}$ is sorted in a descending order of the ranking score. 

We compare ATP with state-of-the-art methods ( {\bf BoW}~\cite{Figueroa2013LRE-BOW}, {\bf Doc2Vec} ~\cite{CQAWord2Vec}, 
{\bf LDA} ~\cite{Ji2012LDA}, {\bf CQARank}~\cite{Yang2013CQArank}), {\bf QDEE}~\cite{QDEE2018}, and {\bf ColdRoute}~\cite{ColdRoute}, based on several popular evaluation criteria such as Mean Reciprocal Rank ({\bf MRR}) ~\cite{Zhu2014}, {\bf Precision$@3$} ~\cite{zhao2017community,ColdRoute}, and {\bf Accuracy} ~\cite{zhao2017community,ColdRoute}. We followed the same settings proposed by Jiankai et al. ~\cite{ColdRoute} to select cold questions for evaluation. 

Table ~\ref{tab:ATP_ColdQuestion_Routing_Table} shows the performance of different approaches on the task of cold question routing, evaluated by MRR,  Precision$@3$ and Accuracy. % Figure~\ref{fig:hge_ColdQuestion_Routing} shows the Accuracy of different approaches on routing cold questions.
 Jiankai et al. ~\cite{ColdRoute} reported that ColdRoute performed consistently better than BoW, Doc2Vec and LDA. To save space, we omitted their performance Table ~\ref{tab:ATP_ColdQuestion_Routing_Table}. 
Based on the results, we can make the following observations:

\begin{itemize}[leftmargin=0.4cm]
    \item ATP performs the best overall evaluation metrics in almost all Stack Exchange sites. For example, ATP improves upon routing metric Accuracy over ColdRoute by $6.14\%$, since ColdRoute fails to take the interactions between questions (asked by the same asker) into consideration.
    ATP improves upon routing metric MRR over QDEE by $7.59\%$, which indicates that incorporating graph reachability and representing user expertise and question difficulty as a feature vector can help ATP identify matching experts for cold questions more accurately and robustly than the state-of-the-art methods. 
    \item ATP performs better than CQARank. The reason is that CQARank's Q$\&$A graph contain more noise than the competition graph used by ATP. The direction of edges in CQARank's Q$\&$A graph is from the asker to the answerer. The underlying assumption is that askers have lower expertise than corresponding answerers. However, Wang et al.~\cite{wang2014} shows that the expertise of the asker is not assumed to be lower than the expertise score of a non-best answerer, since such a user may just happen to see the question and responded that, rather than knowing the answer well. %Take category Python in Stack Overflow for example, it is common to have answers like ``method $x$ provided by user $a$ works for Python 2.7, but I have trouble in running it with Python 3.0''. 
    These kinds of answers do not show corresponding answerers' expertise are higher than the asker's expertise. The generated noise edges in CQARank's Q$\&$A graph can undermine CQARank's performance on experts finding for cold questions.
\end{itemize}

\begin{table}[t!]
\small
\centering
\setlength\tabcolsep{4pt} % default value: 6pt
\caption{Comparisons between ATP and the state of the art methods on cold question routing in CQAs, evaluated by MRR, Precision$@3$ (P$@3$), and Accuracy.}
\begin{tabular}{@{}c|c|ccccc@{}}
 &  & Apple & Gaming & Physics & Scifi & Unix\\ \midrule
%  & BOW & 0.3197 & 0.2908 & 0.343 & 0.2772 & 0.4346\\
%  & Doc2Vec & 0.3481 & 0.2797 & 0.3226 & 0.2979 & 0.4044\\
%  & LDA & 0.3567 & 0.3388 & 0.3956 & 0.3419 & 0.4745\\
 %  & MLP & 0.4271 & 0.374692 & 0.4354 & 0.4153 & 0.4043\\
 MRR & CQARank & 0.4914 & 0.4463 & 0.5315 & 0.4628 & 0.5258\\
 & QDEE & 0.5579 & 	0.6011 & 	0.524 & 	0.5895	 & 0.5158 \\
 & ColdRoute & 0.5365 &	\boldmath\textbf{0.6445} & 0.5288 & \boldmath\textbf{0.6462} &	0.54338 \\
 & \boldmath\textbf{ATP} & \boldmath\textbf{0.574} & 0.6242 & \boldmath\textbf{0.5814} & 0.6405 & \boldmath\textbf{0.5756}\\ \midrule
%  & BOW & 0.0855 & 0.0735 & 0.102 & 0.0466 & 0.2626\\
%  & Doc2Vec & 0.1197 & 0.0415 & 0.0918 & 0.0931 & 0.2323\\
%  & LDA & 0.1197 & 0.0927 & 0.1429 & 0.086 & 0.2862\\
%  P@1 & CQARank & 0.2821 & 0.2268 & 0.2857 & 0.233 & 0.2997\\
%   & QDEE	 & 0.3419  & 	0.3642	 & 0.2755 & 	0.3799 & 	0.2795 \\
%  &  ColdRoute & 0.3291 & 	\boldmath\textbf{0.4505} & 	0.2959 & 	\boldmath\textbf{0.4695} & 	0.3232 \\
%  & \boldmath\textbf{ATP} & \boldmath\textbf{0.3462} & 0.393 & \boldmath\textbf{0.3571} & 0.4373 & \boldmath\textbf{0.367}\\ \midrule
 
%  & BOW & 0.3547 & 0.2812 & 0.3776 & 0.2796 & 0.4444\\
%  & Doc2Vec & 0.3889 & 0.2971 & 0.3571 & 0.276 & 0.4175\\
%  & LDA & 0.4103 & 0.4121 & 0.5204 & 0.4229 & 0.5017\\
%  & MLP & 0.5214 & 0.4441 & 0.5612 & 0.5089 & 0.4949\\
 P@3 & CQARank & 0.5855 & 0.5144 & 0.699 & 0.552 & 0.67\\
 & QDEE &	0.7094 &	0.8019 &	0.6888 &	0.7455 &	0.6566 \\
 & ColdRoute & 	0.6581 & 	0.7796 & 	0.7194 & 	0.7741 & 	0.6869 \\
 & \boldmath\textbf{ATP} & \boldmath\textbf{0.7564} & \boldmath\textbf{0.8179} & \boldmath\textbf{0.7398} & \boldmath\textbf{0.8064} & \boldmath\textbf{0.7205} \\ \midrule 

%  &  BOW	 & 0.3893 & 	0.32 & 	0.4089 & 	0.3302 & 	0.4346 \\
%  & Doc2Vec & 	0.3076 & 	0.3315 & 	0.3641 & 	0.3097	 & 0.4044\\
%  & LDA & 	0.4485 & 	0.4409	 & 0.4946 & 	0.4616 & 	0.4745\\
%   & MLP & 	0.5307	 & 0.4582 & 	0.5283 & 	0.5144	 & 0.4765\\
 Acc. & CQARank & 	0.5555 & 	0.4979 & 	0.6483 & 	0.5693 & 	0.6134\\
 & QDEE  & 	0.6852 & 	0.737 & 	0.6401	 & 0.711 & 	0.6218 \\
 & ColdRoute & 	0.6324 & 	0.7387 & 	0.6354	 & 0.7369 & 	0.6404\\
 & \boldmath\textbf{ATP} & 	\boldmath\textbf{0.7041} & 	\boldmath\textbf{0.7504} & 	\boldmath\textbf{0.6895} & 	\boldmath\textbf{0.7695}	 & \boldmath\textbf{0.6713}\\
  \bottomrule
 \end{tabular}
\label{tab:ATP_ColdQuestion_Routing_Table}
% \vspace{-0.1in}
\end{table}

\section{Conclusion}
\label{sec:conclusion}

In this paper, we have proposed a novel asymmetric transitivity preserving directed graph embedding framework (ATP).
Our scalable embedding technique incorporates both graph hierarchy and reachability information by constructing a novel asymmetric matrix, which is a non-linear transformation of an adjacency matrix (graph reachability) and a diagonal matrix (graph hierarchy).
An efficient factorization based approach is used to generate two embedding vectors for each node to capture the asymmetric transitivity.

With incorporating graph hierarchy and reachability, ATP can perform better than the state-of-the-art in various tasks such as link prediction, and question difficulty estimation and cold question routing in CQAs. And we have proposed several approaches to combine both graph hierarchy and reachability inside a unified framework, and empirically the non-linear transformation works the best.  %Moreover, we believe that our results provide a useful guidance for future designs of CQAs, with respect to routing questions to users with matching expertise and interest. This has great potential to increase question answering quality, satisfaction, helping all the users to improve their expertise (and not just top-level experts), and reduce inefficiency in real-time crowdsourcing of answers. 

As extension of current study, we plan to apply our model to other applications such as community detection in dynamic networks~\cite{wang2018spread} and exception-tolerant abduction ~\cite{Zhang2017AAAI} in attributed networks~\cite{SEANO}. We also would like to address the problem of routing newly posted questions (item cold-start) to newly registered users (user cold-start) in CQAs, with hoping to increase the expertise of the entire community.

 {\bf Acknowledgments} This work is supported by NSF grants CCF-1645599, IIS-1550302, and CNS-1513120, RI xxxxxx, and a grant from the Ohio Supercomputer Center (PAS0166). All content represents the opinion of the authors, which is not necessarily shared or endorsed by their sponsors.

% \newpage
% \clearpage

\bibliographystyle{aaai}
\bibliography{sigproc}

\end{document}